\documentclass[a4paper,twocolumn,10pt]{article}
\usepackage{iccv}
\usepackage{times}
\usepackage{graphicx}
\usepackage{amsmath}
\usepackage{amssymb, bm}
\usepackage[noend]{algpseudocode}
\usepackage{algorithm}
\usepackage{subfigure}
\usepackage{authblk}

\usepackage[pagebackref=true,breaklinks=true,letterpaper=true,colorlinks,bookmarks=false]{hyperref}

\iccvfinalcopy 

\ificcvfinal\fi

\begin{document}

\title{Transformer-based Multi-Instance Learning for Weakly Supervised Object Detection}

\author{
    {Zhaofei WANG$^{1,2}$, Weijia ZHANG$^{3}$, Min-Ling ZHANG$^{1,2}$\thanks{Corresponding to: zhangml@seu.edu.cn.}}\\
    {$^1$ School of Computer Science and Engineering, Southeast University, Nanjing 210096, China}\\
    {$^2$ Key Laboratory of Computer Network and Information Integration 
    (Southeast University), Ministry of Education, China}\\
    {$^3$ School of Information and Physical Sciences, The University of Newcastle, Callaghan, NSW 2308, Australia}\\
    {\tt \small \{wangzf,zhangml\}@seu.edu.cn, \ weijia.zhang@newcastle.edu.au\\}
}
\renewcommand*{\Affilfont}{\small\it} 

\maketitle
\ificcvfinal\thispagestyle{empty}\fi

\begin{abstract}
   Weakly Supervised Object Detection (WSOD) enables the training of object detection models using only image-level annotations.
   State-of-the-art WSOD detectors commonly rely on multi-instance learning (MIL) as the backbone of their detectors and assume that the bounding box proposals of an image are \emph{independent} of each other. 
   However, since such approaches only utilize the highest score proposal and discard the potentially useful information from other proposals, their independent MIL backbone often limits models to salient parts of an object or causes them to detect only one object per class.
   To solve the above problems, we propose a novel backbone for WSOD based on our tailored Vision Transformer named Weakly Supervised Transformer Detection Network (WSTDN). 
   Our algorithm is not only the first to demonstrate that self-attention modules that consider inter-instance relationships are effective backbones for WSOD, but also we introduce a novel bounding box mining method (BBM) integrated with a memory transfer refinement (MTR) procedure to utilize the instance dependencies for facilitating instance refinements. 
   Experimental results on PASCAL VOC2007 and VOC2012 benchmarks demonstrate the effectiveness of our proposed WSTDN and modified instance refinement modules.
\end{abstract}

\section{Introduction}

Convolutional Neural Network (CNN)-based object detection has become a significant area of research in the field of computer vision. Modern object detectors, such as Fast-RCNN \cite{girshick2015fast} and YOLO \cite{redmon2016you}, have achieved remarkable results in Fully Supervised Object Detection (FSOD) tasks, thanks to the availability of large datasets that are labeled with accurate object categories and meticulously annotated bounding boxes. Nevertheless, the process of accurately annotating bounding boxes is laborious and time-consuming, and often incurs significant costs.

To address this challenge, Weakly Supervised Object Detection (WSOD) has emerged as an alternative approach, relying solely on image-level annotations for training. WSOD involves training the detector using only the object category information, while the exact location of the object's bounding box is not required. 
The considerably lower cost of annotation of WSOD has resulted in an increasing amount of research focus in recent years \cite{bilen2016weakly, tang2017multiple, 2018PCL, ren2020instance, seo2022object}.

\begin{figure}[t]
\centering
\begin{center}
\includegraphics[width=0.97\linewidth]{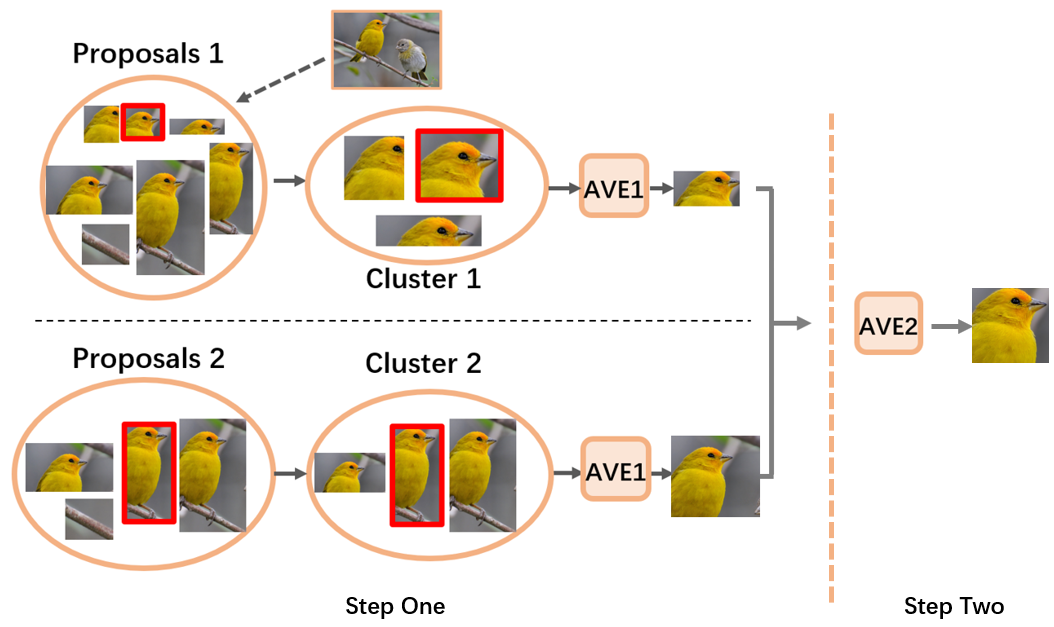}
\end{center}
   \caption{Illustration of the proposed Bounding Box Mining (BBM) module. BBM first considers the spatial relations of proposals by averaging them according to their relevance to the highest score proposals, and then mine novel bounding boxes with different averaging operations. The mined boxes are used as supervision for subsequent modules.
   }
\label{fig1}
\end{figure}

However, the performances of existing WSOD algorithms are still behind those of FSOD methods because its weak supervision signals and algorithm deficiencies.
The performance limitation of existing WSOD algorithms can be attributed to two key factors: the \emph{salient region} problem \cite{shao2022deep} and the \emph{same-class-multi-instance} problem \cite{shao2022deep}. The former issue misleads the model to prioritize the most prominent features of the image, such as the head of a bird or a horse, instead of the entire object. The latter issue occurs when multiple instances of the same class are present, yet the detector fails to identify all of them.

A fundamental cause to the above problem is that most WSOD algorithms builds upon a classic deep multi-instance learning framework, Weakly Supervised Deep Detection Network (WSDDN) \cite{bilen2016weakly}.
WSDDN inherently suffers from both salient-region and same-class-multi-instance problem.   
A large amount of works has been proposed to alleviate the salient region and same-class-multi-instance problems, such as incorporating a refinement module to iteratively alleviate the saliency region \cite{tang2017multiple,zeng2019wsod2} and discovering objects
by measuring the distance between the core instance and its surrounding
boxes \cite{lin2020object,yin2021instance}. 
However, none of the existing methods have touched on the WSDDN backbone, which is one of the root cause for these issues.

As a multi-instance learning framework, the main problem of WSDDN is that it assumes all instances in the same bag to be identically and independently distributed (i.i.d.). 
However, due to the lack of bounding box supervision, the object proposals in WSOD are unlikely to be i.i.d. For instance, the bounding box proposals of an image often contain different parts of an object and overlap. Treating the bounding box proposals as i.i.d. would inevitably mislead the algorithm to either detect only the most prominent part of an object or only one object of several.

In this work, we argue that the crux of the i.i.d. assumption in WSDDN is that it ignores relationships among proposals. 
To solve the problem, we propose to substitute WSDDN with a tailored and novel ViT backbone named Weakly Supervised Transformer Detection Network (WSTDN). 
To the best of our knowledge, WSTDN is the first WSOD algorithm with a ViT backbone, which exploits the multi-head self-attention module and is capable for handling the semantic relationships among proposals.

Furthermore, we propose a Bounding Box Mining (BBM) module and Memory Transfer Refinement (MTR) for modeling the spatial relationships among proposals and escaping saliency regions. As shown in Figure~\ref{fig1}, BBM first forms different clusters of potential ground-truth boxes based on their spatial relationships, and then synthesizes new boxes using a adaptive coordinate averaging procedure.  
Then, MTR is appled before BBM-based refinements to iteratively improve the effectiveness of the outputs using momentum.

We summarize our contributions as follows:
\begin{enumerate}
    \item We empirically demonstrate that naive implementation of self-attention does not work for WSOD and propose a novel WSTDN backbone based on a tailored ViT for exploiting the relationships among proposals. 
    \item We propose the Bounding Box Mining (BBM) procedure which is capable of synthesizing new proposals by incorporating spatial information of existing boxes. 
    \item We propose the Memory Transfer Refinement(MTR) mechanism which helps enhancing iterative BBM and avoiding saliency region traps.
\end{enumerate}

\section{Related Work}
\subsection{Transformers} 
Transformer \cite{vaswani2017attention} has garnered significant attention in natural language processing. The crux of Transformers is the multi-head self-attention mechanism \cite{vaswani2017attention}, which has the capability to approximate CNN \cite{cordonnier2019relationship}.
Following Vaswani \etal's \cite{vaswani2017attention} pioneering work, Vision Transformer (ViT) \cite{dosovitskiy2020image} extends Transformer to computer vision tasks. ViT divides images into patches, mimicking the tokens in the original Transformer, while abandoning Transformer Encoders. Recently, ViT has increasingly received attention and have been applied to tasks such as fully supervised object detection and semantic segmentation. For instance, Swin Transformer \cite{liu2022video} utilizes a local self-attention mechanism to improve the ability to extract local features. DETR \cite{carion2020end} extends ViT to fully supervised objection detection and achieves comparable accuracy and speed to Fast-RCNN \cite{girshick2015fast}.

\subsection{Deep Multi-Instance Learning}

A widely adopted way for dealing with WSOD is the Multi-Instance Learning (MIL) framework. MIL deals with training data organized in sets (called bags) where each bag may contain various numbers of samples (called instances). During training, MIL algorithms only have access to the labelsof the bags, without knowing the labels of instances. 
Deep MIL algorithms using fixed \cite{wang2018revisiting} or attention-based pooling mechanisms \cite{ilse2018attention} have attracted increasing attentions; however, most of the deep MIL algorithms treat instances as i.i.d.
Notably, some recent deep MIL algorithms accommodates semantic relationships among instances using self-attention \cite{rymarczyk2021kernel} and deep generative modeling \cite{zhang2021mivae,zhang2022cmil}. Unfortunately, none of them can be applied to the problem characteristics of WSOD.

\begin{figure*}[t]
\begin{center}
\includegraphics[width=0.87\linewidth]{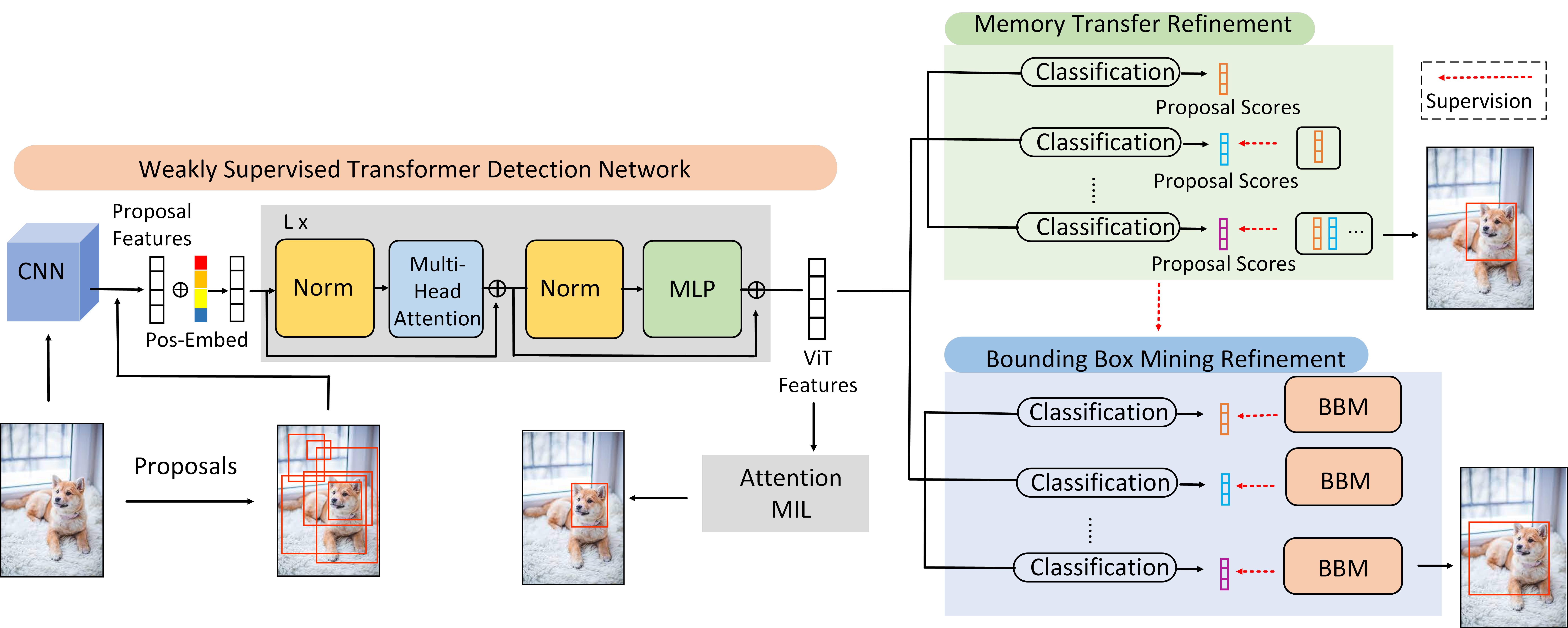}
\end{center}
   \caption{Overview of the proposed WSOD architecture. The left component is our proposed WSTDN backbone, and the right components are the Memory Transfer Refinements (MTR) and Bounding-Box-Mining (BBM) based proposal refinements. 
   The output of WSTDN serves as the supervision for MTR, while the output of MTR serves as the supervision for BBM, which generates the final results.}
\label{fig2}
\end{figure*}

\subsection{Weakly Supervised Object Detection}
The initial step in the majority of WSOD algorithms consists a proposal generation procedure. Since the proposals of an image can be regarded as a bag of instances under the MIL framework, Weakly Supervised Deep Detection Network (WSDDN) \cite{bilen2016weakly} is one of the first WSOD algorithms that applies two MIL classification branches. Many of the later WSOD methods use WSDDN as their backbones. 
Tang \etal \cite{tang2017multiple} proposed Online Instance Classifier Refinement (OICR) for iteratively refining the results of WSDDN. Ren \etal \cite{ren2020instance} showed that bounding box regression is an useful addition to instance refinements. 

For combating the salient region problem, Ren \etal\cite{ren2020instance} proposed to drop pixels from salient regions for facilitating detection, and Huang \etal \cite{huang2020comprehensive} proposed to combine the attention maps of different image augmentations and feature maps of different CNN layers. For the same-class-multi-instance problem, Lin \etal \cite{lin2020object} proposed Online instance Mining (OIM) using appearance graph and spatial graph, and Gao \etal \cite{gao2019c} proposed to localize multiple objects with coupled multi-instance networks. Furthermore, some recent WSOD algorithms address the weak supervision using noisy label learning \cite{huang2022w2n} and integrating fully-supervised/semi-supervised learning \cite{sui2022salvage}.

Unfortunately, to the best of our knowledge, none of the existing WSOD algorithms tackles saliency region and same-class-multi-instance problem by considering the relationships among instances with the self-attention mechanism, nor have they addressed the root cause of the problem inherited by the WSDDN backbone.

\section{The Proposed Method}

In this section, we discuss the details of the three main components of our algorithm with an emphasize on both their individual design characteristics and collaborated synergies.

\subsection{Overall Framework}

Our overall network framework is shown in Figure~\ref{fig2}. The WSOD dataset comprises an input image accompanied by its label $Y = \{y_1, y_2, ..., y_C\} \in \mathbb{N}^C$, where $C$ denotes the number of categories. In contrast to the FSOD datasets, no bounding boxes are provided during training.
The image is fed into our ViT-based WSTDN module for classification, which accommodates relationships among proposals and generates the pseudo ground-truth data utilized for supervising the MTR module. 
The MTR module utilizes the spatial relationship from WSTDN and iteratively refine instances to improve bounding boxes. Finally, the output of MTR is used as the input to supervise Bounding Box Mining (BBM) for further refinements.

\subsection{Weakly Supervised Transformer Detection Net}

The image regions cropped by the bounding box proposals naturally align with ViT's idea of splitting an image into patches.
Specifically, the image and a total of $N$ candidate proposals extracted by Selective Search (SS) \cite{uijlings2013selective} are fed into a pre-trained CNN using ImageNet dataset\cite{deng2009imagenet}. 
It is worth noting that the pre-trained CNN feature extractor of WSTDN can be substituted with a full transformer-based feature extractor trained from scratch. 
As shown in a standard ViT, it is preferred to train a CNN feature extractor when the data is limited and train a full transformer model when both data and computing resources are abundant \cite{dosovitskiy2020image}. In our work, to ensure a fair comparison with existing WSOD algorithms, we use the hybrid architecture without introducing additional data.
The extracted features are then fed into the ROI Pooling layers to generate features of the same size, which is flattened by two fully-connected layers. In the rest of the paper, we denote the flattened features as proposal feature vectors $P = \{\bm{p}_1, \bm{p}_2, ..., \bm{p}_N\}$ and their corresponding bounding boxes generated by selective search as $B = \{\bm{b}_1, \bm{b}_2, ..., \bm{b}_N\}$. It is evident that the proposal feature vectors $P$ correspond to the embedding tokens of ViT.

In order to exploit the relationships among proposals, we utilize the position embedding in ViT and add it to the proposal feature vectors. The position embedding can be directly added to the proposal features as they have the same dimension, and the embedding elements can be learned by back propagation. 
However, different from a standard ViT, we substitute the ViT classification head with an averaging operation, since averaging the proposals can better utilize their information than a single classification head, especially when annotations are incomplete in WSOD.

As illustrated by the grey shaded component of WSTDN in Figure~\ref{fig2}, the proposal feature vectors and their position embeddings are then fed into ViT encoders, whose iterative multi-head-self-attention module further models the semantic relationships among proposals. 
The ViT encoders then output the ViT feature vector $V = \{\bm{v}_1, \bm{v}_2, ..., \bm{v}_N\}$, which is used as the input of the attention-based MIL module \cite{ilse2018attention}.

Our attention-based MIL module has a classification branch and an attention branch, which collaboratively generate the proposal scores matrix $X$ of size $R^{C \times N}$. 
Then, $p_c = \sum_{n=1}^Nx_{cn}$ is aggregated as the image-level prediction where $x_{cn}$ is the element of the proposal score matrix $X$. Finally, the aggregated $p_c$ is used to calculate the multi-class cross entropy loss $L_{CE}$:

\begin{equation}
    L_{CE} = -\sum_{c=1}^Cy_c\log p_c + (1-y_c)\log(1-p_c)
\end{equation}

It is worth noting that a straightforward implementation of the self-attention mechanism, e.g., a self-attention based method for general MIL tasks \cite{rymarczyk2021kernel}, will not work for WSOD. We will further discuss its reasons and empirically demonstrate this in the experiment section.

\begin{figure}[t]
\begin{center}
\subfigure[]{
\includegraphics[width=0.47\linewidth]{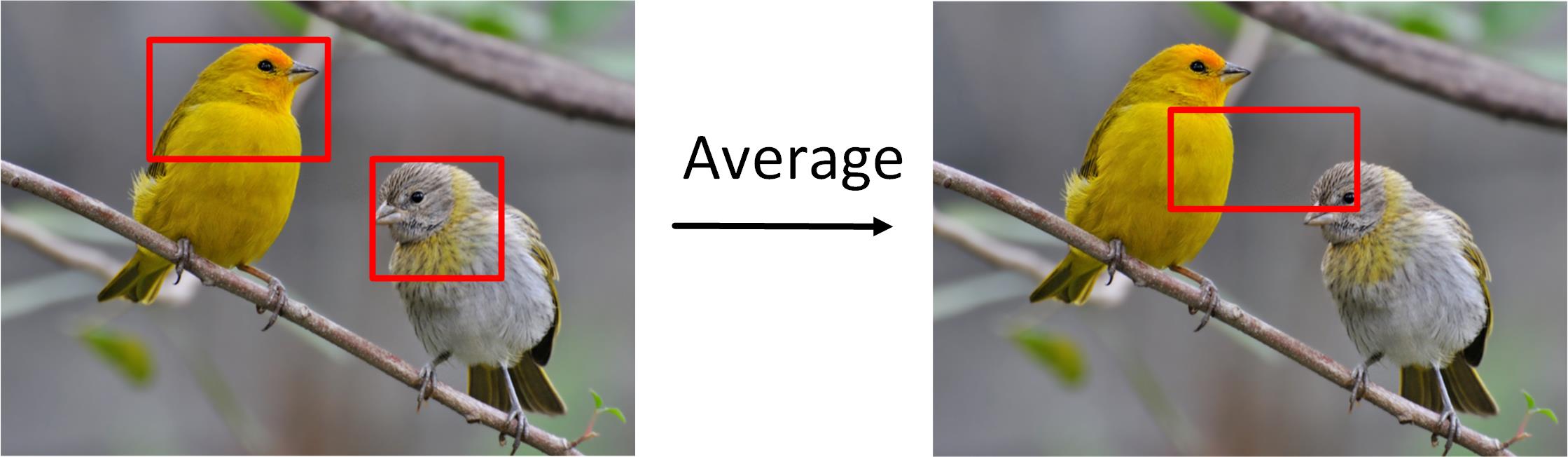}
}
\hspace{0.4cm}
\subfigure[]{
\includegraphics[width=0.4\linewidth]{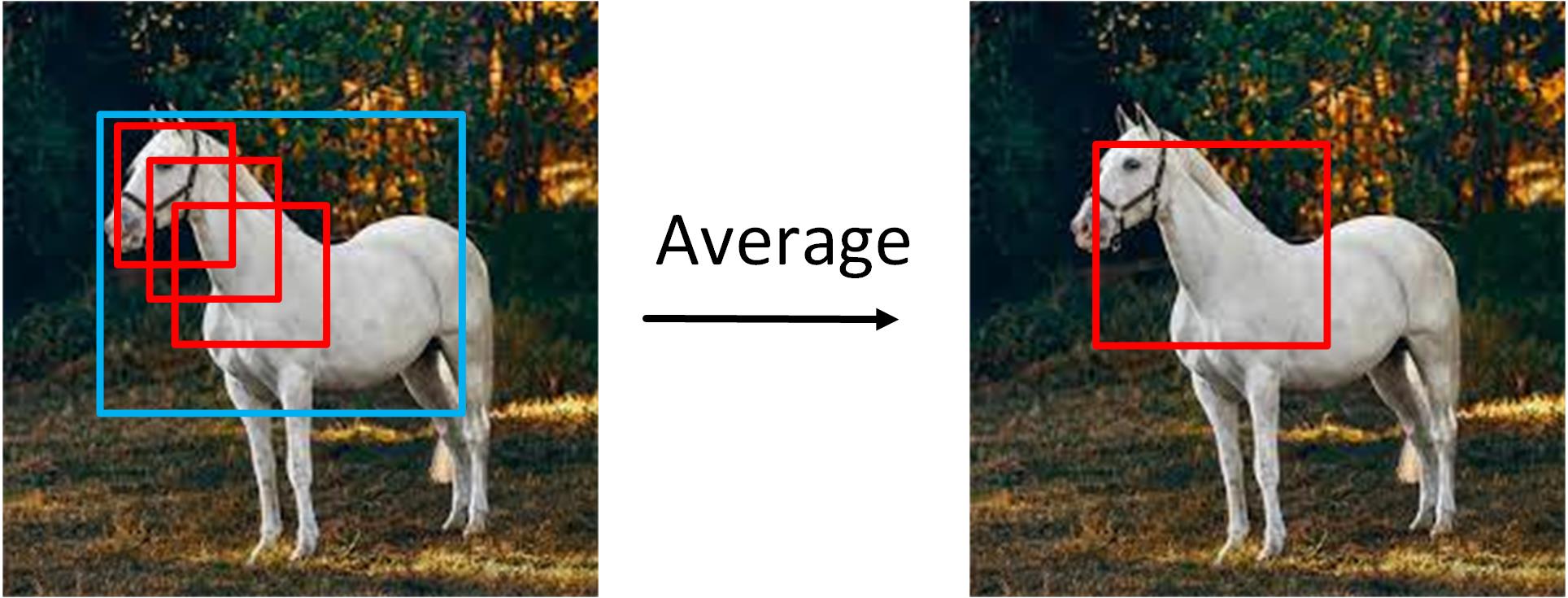}
}
\end{center}
   \caption{Invalid averaging operation caused by (a) same-class-multi-instance and (b) boxes containing smaller saliency parts. The blue box in (b) indicates the box containing larger parts.}
\label{fig3}
\end{figure}

\subsection{Bounding Box Mining}

In most existing WSOD methods, the highest score proposal is selected and used as the supervision for instance refinements; however, this approach often stucks in salient regions because it ignores the potentially useful information of other high score proposals. 
To address this problem and further utilize the relationship among proposals, we propose the Bounding Box Mining (BBM) to complement our proposed WSTDN backbone.

The overall process of BBM is shown in Algorithm 1 which is consisted of two main steps.
In the first step, we select the highest score proposal $\bm{p}_{max-1}$ from proposals set $P_1 : = P = \{\bm{p}_1, \bm{p}_2, ..., \bm{p}_N\}$. 
Then, proposals in $P_1$ are grouped into cluster1 $C_1 = \{\bm{p}_{11}, \bm{p}_{12}, ..., \bm{p}_{1|C1|}\}$ if their overlaps with $\bm{p}_{max-1}$ are greater than $\gamma_{1}$ when measured with IOU. 
After that, a coordinate averaging operation is applied to the bounding boxes belonging to $C_1$, creating a novel bounding box which we denote as $\bm{b}_{create-1}$. 
Then, $C_1$ is removed from $P_1$ to obtain a new proposal set $P_2$ where we select the top-1 score proposal $\bm{p}_{max-2}$ from $P_2$. Following the same procedure we obtain the second cluster $C_2 = \{\bm{p}_{21}, \bm{p}_{22}, ..., \bm{p}_{2|C2|}\}$ and $\bm{b}_{create-2}$. 
This iterative procedure of generating novel bounding boxes is repeated until we reach the pre-defined $Q$-th box $\bm{b}_{create-Q}$. 

The iterative box averaging operation creates flexible and novel bounding boxes which integrate information from different proposals.
Furthermore, since the proposals are iteratively clustered, different sets of proposals can find bounding boxes containing object parts other than the salient ones. 
As shown in Figure~\ref{fig1}, proposals in the second cluster contain more object parts than their first counterpart. However, the generated boxes may contain backgrounds and different object parts. Therefore, a further combination procedure is necessary.

 \begin{algorithm}[t]
 \caption{Bounding Box Mining (BBM)}
 \hspace*{0.02in}{\bf Input:}
 Proposals $P$, corresponding boxes $B$ \\
 \hspace*{0.02in}{\bf Output:}
 supervision box $\bm{b}_{sup}$
 \begin{algorithmic}[1]
 \State STEP 1:
 \For{$i$ = 1 to $Q$}
    \State $C_i = \emptyset$
    \State Select $\bm{p}_{max-i}$ from $P$
    \For {$\bm{p}_{j}$ in $P$}
        \If {IOU$(\bm{p}_{j}, \bm{p}_{max-i}) > \gamma_{1}$}
            \State put $\bm{p}_{j}$ into $C_i$
        \EndIf
    \EndFor
    \State $\bm{b}_{create-i} \leftarrow$ Average boxes corresponding to $C_i$
    \State Delete $C_i$ from $P$
 \EndFor
 \\
 \State STEP2:
 \State $B_{create} \leftarrow \{\bm{b}_{create-1}, \bm{b}_{create-2}, ..., \bm{b}_{create-Q}\}$
 \For{$i$ = 1 to $Q$}
    \State Delete $\bm{b}_{create-i}$ containing multi-instance
    \State Enhance $\bm{b}_{create-i}$ containing larger parts
\EndFor 
\State $\bm{b}_{sup} \leftarrow$ Average $B_{create}$
\State $\bm{b}_{sup} \leftarrow$ Average ($\bm{b}_{sup}$, $\bm{b}_{max-i}$)
 \end{algorithmic}
 \end{algorithm}

Therefore, the second step of BBM need to focus on reducing the background noises and combining different object parts, whose overall illustration is provided in the Average2 (AVE2) operation in Figure~\ref{fig1}. Denoting the set of created boxes in the first step as $B_{create} = \{\bm{b}_{create-1}, \bm{b}_{create-2}, ..., \bm{b}_{create-Q}\}$, there are several design considerations when averaging the coordinates of bounding boxes in $B_{create}$ in AVE2.

First, the synthesized boxes in $B_{create}$ may contain multiple instances of the same class, e.g., they contain multiple birds as illustrated in Figure~\ref{fig1}. 
In this circumstance, the average operation would be inappropriate, as shown in Figure~\ref{fig3}(a). 
In order to solve this issue, if the bounding box $\bm{b}_{create-i}$ does not overlap with $\bm{b}_{max-1}$, then it will be eliminated from the set $B_{create}$. Here, $\bm{b}_{max-1}$ refers to the corresponding bounding box of the proposal $\bm{p}_{max-1}$. 
The second issue pertains to the possibility of bounding boxes that contain larger parts being dominated by those that contain smaller parts, as depicted in Figure~\ref{fig3}(b), after the averaging operation. To circumvent this problem, we propose an adaptive approach to assign weights to the bounding boxes based on their size during the averaging operation. Specifically, boxes containing larger parts will be assigned relatively higher weights.

\subsection{BBM-based Instance Refinement}
By leveraging the proposed bounding box mining technique, we combine it with iterative instance refinements, as depicted in the blue block on the right-hand side of Figure~\ref{fig2}, to further improve detection from weak supervision.

The refinement process consists of $K$ stages of branches, where each branch utilizes the ViT features as input and produces a proposal scores matrix $X^{k}$. This matrix corresponds to the classification vectors of all proposals. To generate supervision for the initial refinement, BBM is applied to the proposal scores matrix of the last refinement stage of MTR, which yields the supervision $\bm{b}_{sup}^1$. Subsequently, for the $k$-th refinement, where 2 $\leq k \leq K$, BBM is applied to the proposal scores matrix of the ($k-$1)-th refinement, resulting in the supervision $\bm{b}_{sup}^k$.

The loss function of BBM-based instance refinements is similar as OICR proposed by Tang \etal \cite{tang2017multiple}. In $k$-th refinement, suppose the target category is class $c$, we label the $n$-th proposal to class $c$, $i.e.$ $y_{cn}^{k}=1$, if its IOU with  $\bm{b}_{sup}^k$ is greater than $\gamma_{2}$; otherwise, we label it to background, $i.e.$ $ y_{(C+1)n}^{k}=1$. The loss function is as Eq. 2. 
Also, we add a bounding-box regression layer parallel to the last refinement branch, the loss of the regression layer is the Smooth L1 loss\cite{girshick2015fast} commonly used in FSOD.  

\begin{equation}
    L_{BBM} = -\frac{1}{|N|}\sum_{k=1}^{K}\sum_{n=1}^{N}\sum_{c=1}^{C+1}y_{cn}^{k}logx_{cn}^{k}
\end{equation}

It is worth noting that in order to achieve improved and consistent detection results in the $k$-th refinement, we employ both the bounding boxes generated by BBM and the initial highest score proposals generated by the ($k-$1)-th refinement as supervision.

\subsection{Memory Transfer Refinement}
The final component to our proposed model is the iterative Memory Transfer Refinement (MTR) module, which is inspired by the importance of historical data in time series analysis. 
In many fields, the analysis of time series data often involves the consideration of historical data owing to the rich information it provides for current data. For instance, momentum gradient descent \cite{liu2020improved}, temporal difference method \cite{menache2005basis} in reinforcement learning, and high-order Markov Chains all utilize historical data for modeling.

As illustrated in the right green block of Figure~\ref{fig2}, our MTR module regards proposal scores of all stages of refinements as a time series $T = \{S_1, S_2, ..., S_k\}$, where $S_k$ ($1 \leq k \leq K$) indicates proposal scores in the $k$-th refinement. The proposal scores for supervising the $k$-th refinement ${Sup}_k$ ($2 \leq k \leq K$) is calculated according to Eq. 3 and Eq. 4. ${Sup}_1$ is the score matrix $X$ from WSTDN.

\begin{equation}
\begin{split}
    Sup_k = \sum_{i=1}^{k-1} \frac{\alpha_i}{k - 1} S_i \\
\end{split}
\end{equation}

\begin{equation}
    \alpha_i = \left\{
    \begin{array}{rcl}
    1 + \frac{(k - 1)(k - 2)}{2} \delta & \mbox{for} & i=k-1 \\[5pt]
    1 - (k - i - 1) \delta & \mbox{for} & i \neq k-1 \\
    \end{array}
    \right.
\end{equation}

In Equation 3, $\alpha_i / (k-1)$ indicates weights assigned to $S_i$. Furthermore, the weights satisfy two requirements: 1) all the weights add up to one; 2) if $i$ is closer to $k$, the weight of $S_i$ is larger. The second requirement is based on the assumption that data closer to the current time is more reliable. The loss function of MTR $L_{MTR}$ is the same as BBM-based refinements.

\subsection{Method Summary}

To sum up, the overall loss function of our method can be described in Eq. 5:

\begin{equation}
    L_{Total} = L_{CE} + L_{MTR} + L_{BBM} 
\end{equation}

Since VIT is based on the self-attention mechanism, our method can be regarded as a deep MIL method that integrates self-attention with attention mechanism for considering the relationships among instances. To the best of our knowledge, this is the first deep MIL methods that utilizes self-attention and is tailored for the WSOD task. Previous MIL algorithms, e.g., \cite{rymarczyk2021kernel} and \cite{zhang2021mivae}, are designed for general MIL classification tasks and are thus not suitable for WSOD.

\begin{table*}[t]
\begin{center}
\scriptsize
\tabcolsep=0.065cm
    \begin{tabular}{c|c|cccccccccccccccccccc|c}
        \hline
        Backbone& Method &Aero& Bike& Bird& Boat& Bottle& Bus& Car& Cat& Chair& Cow& Table& Dog& Horse& Motor& Person& Plant& Sheep& Sofa& Train& TV &mAP\\
        \hline \hline
        &WSDDN\cite{bilen2016weakly}&39.4	&50.1&	31.5	&16.3	&12.6	&64.5&	42.8	&42.6&	10.1&	35.7&	24.9&	38.2&	24.4&	55.6&	9.4&	14.7&	30.2&	40.7&	54.7&	46.9& 34.8 \\
        &OICR\cite{tang2017multiple} &58.0	&62.4&	31.1&	19.4&	13.0&	65.1&	62.2&	28.4&	14.8&	44.7&	30.6&	25.3&	37.8&	65.5&	15.7&	24.1&	41.7&	46.9&	64.3&	62.6& 41.2 \\
        &MELM\cite{wan2018min} &55.6 &66.9& 34.2& 29.1& 16.4& 68.8& 68.1& 43.0& 25.0& 65.6& 45.3& 53.2& 49.6& 68.6& 2.0& 25.4& 52.5& 56.8& 62.1& 57.1& 47.3 \\
        &WSRPN\cite{tang2018weakly} & 60.3& 66.2& 45.0& 19.6& 26.6& 68.1& 68.4& 49.4& 8.0& 56.9& 55.0& 33.6& 62.5& 68.2& 20.6& \textbf{29.0}& 49.0& 54.1& 58.8& 58.4& 47.9 \\
        &C-MIL\cite{wan2019c} & 62.5 &58.4 &49.5 &32.1 &19.8 &70.5 &66.1 &63.4& 20.0 &60.5& 52.9& 53.5& 57.4& 68.9& 8.4 &24.6& 51.8& 58.7& 66.7& 63.5 & 50.5 \\
        WSDDN\cite{bilen2016weakly}&WSOD2\cite{zeng2019wsod2} & 65.1 &64.8& 57.2& \textbf{39.2}& 24.3& 69.8& 66.2& 61.0& 29.8& 64.6& 42.5& 60.1& \textbf{71.2}& \textbf{70.7}& 21.9& 28.1& \textbf{58.6}& 59.7& 52.2& 64.8 &53.6\\
        &C-MIDN\cite{gao2019c} & 53.3&71.5& 49.8& 26.1& 20.3& 70.3& 69.9& 68.3& 28.7& 65.3& 45.1& 64.6& 58.0& 71.2& 20.0& 27.5& 54.9& 54.9& 69.4& 63.5 & 52.6\\
        &OIM\cite{lin2020object} & 55.6 &67.0& 45.8& 27.9& 21.1& 69.0& 68.3& 70.5& 21.3& 60.2& 40.3& 54.5& 56.5& 70.1& 12.5& 25.0& 52.9& 55.2& 65.0& 63.7& 50.1 \\
        &SLV\cite{chen2020slv} &\textbf{65.6}& 71.4& 49.0& 37.1& 24.6& 69.6& 70.3& 70.6& \textbf{30.8}& 63.1& 36.0& 61.4& 65.3& 68.4& 12.4& 29.9& 52.4& 60.0& 67.6& 64.5& 53.5 \\
        \hline 
        & OICR+FRCNN\cite{tang2017multiple} & 65.5& 67.2& 47.2& 21.6& 22.1& 68.0& 68.5& 35.9& 5.7& 63.1& 49.5& 30.3& 64.7& 66.1& 13.0& 25.6& 50.0& 57.1& 60.2& 59.0& 47.0 \\
        WSDDN\cite{bilen2016weakly}&C-MIL+FRCNN\cite{wan2019c} & 61.8& 60.9& 56.2& 28.9& 18.9& 68.2& 69.6& 71.4& 18.5& 64.3& \textbf{57.2}& \textbf{66.9}& 65.9&65.7& 13.8& 22.9& 54.1& \textbf{61.9}& 68.2& \textbf{66.1}& 53.1 \\
        &OIM+FRCNN\cite{lin2020object} & 53.4& 72.0& 51.4& 26.0& 27.7& 69.8& 69.7& 74.8& 21.4& \textbf{67.1}& 45.7& 63.7& 63.7& 67.4& 10.9& 25.3& 53.5 &60.4& \textbf{70.8}& 58.1& 52.6 \\
        \hline
        WSTDN&Ours(single)&60.1 &\textbf{76.8} &\textbf{59.9} &31.7  &\textbf{29.9} &\textbf{73.4} &\textbf{72.0} & \textbf{74.9} &29.2 &64.4 &47.3 &41.0 &61.6 &69.1 &\textbf{33.7} &25.0 &57.3 &61.4 &67.3 &58.0 &\textbf{54.7} \\
        \hline
    \end{tabular}
\end{center}
\caption{Comparison with the state-of-the-arts in terms of mAP (\%) on VOC2007 $test$ set.}
\label{tab1}
\end{table*}

\begin{table*}[t]
\begin{center}
\scriptsize
\tabcolsep=0.065cm
    \begin{tabular}{c|c|cccccccccccccccccccc|c}
        \hline
        Backbone&Method &Aero& Bike& Bird& Boat& Bottle& Bus& Car& Cat& Chair& Cow& Table& Dog& Horse& Motor& Person& Plant& Sheep& Sofa& Train& TV &mean\\
        \hline \hline
        &WSDDN\cite{bilen2016weakly}&65.1 &58.8& 58.5& 33.1& 39.8 &68.3& 60.2& 59.6& 34.8& 64.5& 30.5& 43.0& 56.8& 82.4& 25.5& 41.6& 61.5& 55.9& 65.9&3.7& 53.5 \\
        &OICR\cite{tang2017multiple} &81.7& 80.4& 48.7& 49.5& 32.8& 81.7& 85.4& 40.1& 40.6& 79.5& 35.7& 33.7& 60.5& 88.8& 21.8& 57.9& 76.3& 59.9& 75.3& 81.4& 60.6\\
        &MELM\cite{wan2018min} & - &-&-&-&-&-&-&-&-&-&-&-&-&-&-&-&-&-&-&-& 61.4 \\
        &WSRPN\cite{tang2018weakly} & 77.5 &81.2& 55.3& 19.7& 44.3& 80.2& 86.6& 69.5& 10.1& \textbf{87.7}& \textbf{68.4}& 52.1& 84.4& 91.6& \textbf{57.4}& \textbf{63.4}& 77.3& 58.1 &57.0 &53.8& 63.8 \\
        &C-MIL\cite{wan2019c} & - &-&-&-&-&-&-&-&-&-&-&-&-&-&-&-&-&-&-&-& 65.0 \\
        WSDDN\cite{bilen2016weakly}&WSOD2\cite{zeng2019wsod2}  & 87.1 &80.0& \textbf{74.8}& \textbf{60.1}& 36.6& 79.2& 83.8& 70.6& 43.5& 88.4& 46.0& 74.7& 87.4& 90.8& 44.2& 52.4& 81.4& 61.8& 67.7& 79.9& 69.5 \\
        &C-MIDN\cite{gao2019c} &  - &-&-&-&-&-&-&-&-&-&-&-&-&-&-&-&-&-&-&-& 68.7 \\
        &OIM\cite{lin2020object} &  - &-&-&-&-&-&-&-&-&-&-&-&-&-&-&-&-&-&-&-& 67.2 \\
        &SLV\cite{chen2020slv} &84.6 &84.3 &73.3& 58.5& 49.2& 80.2& 87.0& \textbf{79.4}& \textbf{46.8}& 83.6& 41.8& \textbf{79.3}& \textbf{88.8}& 90.4& 19.5& 59.7& 79.4 &\textbf{67.7} &\textbf{82.9}& \textbf{83.2}& \textbf{71.0} \\
        \hline         WSDDN\cite{bilen2016weakly}&OICR+FRCNN\cite{tang2017multiple} & 85.8& 82.7& 62.8& 45.2 &43.5& \textbf{84.8} &87.0& 46.8& 15.7& 82.2& 51.0& 45.6 &83.7& 91.2& 22.2& 59.7& 75.3& 65.1& 76.8& 78.1& 64.3 \\
        &OIM+FRCNN\cite{lin2020object} & - &-&-&-&-&-&-&-&-&-&-&-&-&-&-&-&-&-&-&-& 68.8 \\
        \hline        WSTDN&Ours(single)&\textbf{86.3}&\textbf{85.9}&72.7&52.1 &\textbf{52.7} &82.7 &\textbf{87.8} & 79.1 &45.5 &83.6 &46.0 &52.3 &81.0 &\textbf{91.6} &51.4 &58.2 &\textbf{86.6} &62.1 &80.2 &70.6 &70.4 \\
        \hline
    \end{tabular}
\end{center}
\caption{Comparison with the state-of-the-arts in terms of CorLoc (\%) on VOC2007 $trainval$ set.}
\label{tab2}
\end{table*}

\section{Experiments}
\subsection{Datasets and Evaluation Metrics}

Our method is evaluated on the popular PASCAL VOC2007 and VOC2012 benchmark datasets \cite{everingham2009pascal} as used by previous state-of-the-art WSOD algorithms. 
These datasets are composed of 20 categories and contains 9,962 and 22,531 images, respectively. 
However, in contrast to fully supervised object dection, WSOD algorithms only have access to image-level annotations.

The datasets are split into trainval set and test set. For both datasets, the trainval sets (which contain 5,011 and 11,540 images respectively) are used to train our network, and the test set (4,952 and 10,991 images respectively) are used for evaluation. 
All algorithms are evaluated with two widely used metrics: mAP and CorLoc. The mean of AP (mAP) is used to evaluate our model on test set, following the standard PASCAL VOC protocols. Correct Localization (CorLoc) \cite{deselaers2012weakly} is used to evaluate the models bounding box localization capability on trainval set. 
A bounding box is considered positive if it has an $IoU >$ 0.5 with a ground-truth bounding box.

\begin{figure*}[t]
\begin{center}
\includegraphics[width=0.8\linewidth]{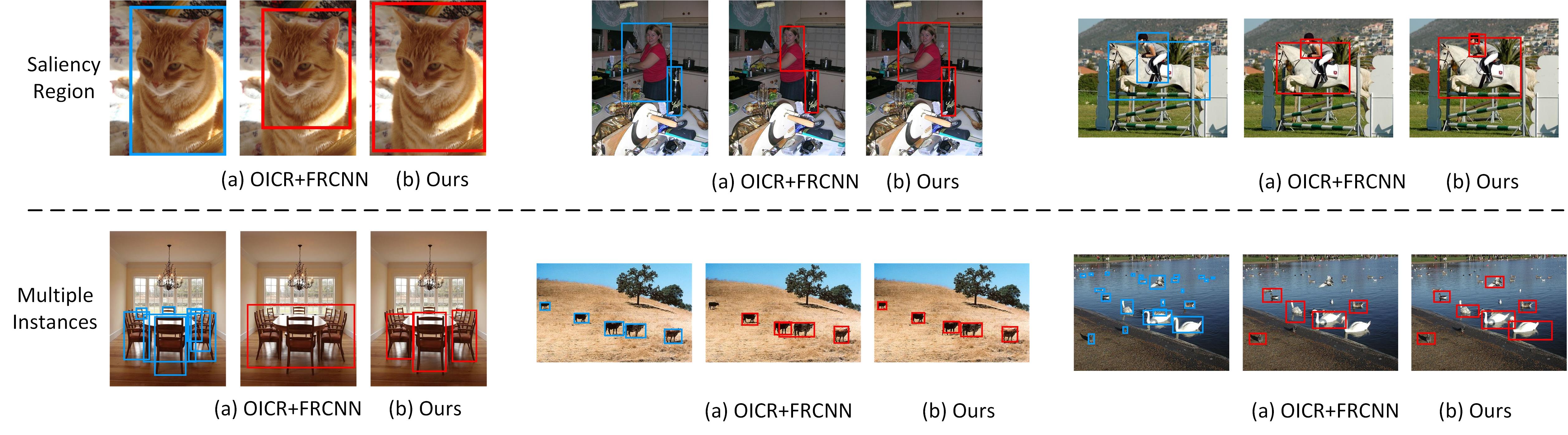}
\end{center}
   \caption{Comparison of results generated by (a) OICR + FRCNN, (b) ours on VOC2007 $test$ set. Blue boxes indicate groundtruths and red boxes indicate detection results. Results with scores $\geq$ 0.15 are shown. The first line illustrates the effectiveness of our method on solving salient region problem. The second line illustrates the effectiveness of our method on solving same-class-multi-instance problem.}
\label{fig4}
\end{figure*}

\subsection{Implementation Details}

Following the same setting as previous WSOD methods, we utilize VGG16 model pre-trained on the Imagenet dataset \cite{deng2009imagenet} as the feature extractor of our backbone network. 
The usage of a pre-trained VGG16 feature extractor is to ensure a fair comparison that our method uses the same amount of training data as other WSDDN-based algorithms. If more training data is available, WSTDN can also be trained from scratch with ViT feature extractors.
The last max-pooling layer and subsequent convolution layers are replaced by the dilated convolution layers, following Tang \etal \cite{tang2017multiple}. Selective Search \cite{uijlings2013selective} is used to generate proposals for datasets. During training, the mini-batch size is set to 2, and the learing rate is set to 0.001 and 0.0001 for the first 60K iterations and the following 30K iterations, respectively.The momentum is set to 0.9. The IOU threshold $\gamma_{1}$ and $\gamma_{2}$ in BBM are set to 0.3 and 0.5 respectively. The number of Transformer Encoders is set to 2. The high score proposals are kept and Non-Maximum Suppression (with 30 \% IOU threshold) is applied to calculte mAP and CorLoc.
Our experiments are conducted using an NVIDIA GeForce RTX 3090 GPU with 24GB of memory.

\subsection{Comparison with State-of-the-arts}
In this section, we compare our method with recently proposed state-of-the-arts WSOD algorithms on both PASCAL VOC 2007 and VOC 2012 benchmarks. 
The detail results for VOC 2007 are shown in Table~\ref{tab1} and Table~\ref{tab2}, where FRCNN indicates that the detection results of WSOD algorithms are used as pseudo ground-truth and fed into a Fast-RCNN detector \cite{girshick2015fast} for further training.

From Table~\ref{tab1} we can see that our method achieves the highest mAP of 54.7 \%. Furtheremore, the results in Table~\ref{tab2} show that our method achieves 70.4 \% of CarLoc and outperforms all the other standalone methods and methods that combine FRCNN, except for SLV \cite{chen2020slv}. 
Specifically, our method achieves state-of-the-arts detection results on category \emph{Bike, Bird, Bottle, Bus, Car, Cat, Person}. Relatively high detection results are also seen on category \emph{Chair, Cow, Motor, Sheep}. 
It can be obtained that our method performs well in detecting rigid objects such as bottles and buses, and also simple non-rigid objects such as cats and birds, thanks to the semantic and spatial relations mined by WSTDN and BBMs.
However, all methods do not perform well on complex non-rigid objects, e.g., \emph{Person}. Surprisingly, we achieve the highest detection result on \emph{Person}, but it is still much lower than other categories. It indicates that accurate detection of non-rigid objects is still a fundamental challenge in WSOD.
Last, Table~\ref{tab3} shows the CorLoc results of our method and baselines on PASCAL VOC 2012. 


\begin{table}[t]
\begin{center}
\small
    \begin{tabular}{|c|c|c|c|}
        \hline
        Baseline&Method & CorLoc\\
        \hline
        &OICR\cite{tang2017multiple} & 62.1\\
        &MELM\cite{wan2018min}  &- \\
        &WSRPN\cite{tang2018weakly}  &64.9 \\
        &C-MIL\cite{wan2019c}  &67.4 \\
        WSDDN\cite{bilen2016weakly}&WSOD2\cite{zeng2019wsod2}  &\textbf{71.9} \\
        &C-MIDN\cite{gao2019c} &71.2\\
        &OIM\cite{lin2020object} &67.1 \\
        &SLV\cite{chen2020slv} &69.2 \\
        \hline 
        WSDDN\cite{bilen2016weakly}&OICR+FRCNN\cite{tang2017multiple} &- \\
        &OIM+FRCNN\cite{lin2020object} &69.5\\
        \hline
        WSTDN&Ours(single)&70.7 \\
        \hline
    \end{tabular}
\end{center}
\caption{Comparison with the state-of-the-arts in terms of CorLoc (\%) on VOC2012 $trainval$ set.}
\label{tab3}
\end{table}

We present the qualitative detection results in Figure~\ref{fig4}, where the detection results of six different images are arranged in six clusters. 
In each cluster, the left image indicates the ground-truth bounding box, the middle image indicates the boxes generated by OICR + FRCNN\cite{tang2017multiple}, and the right image indicates boxes generated by our method. 
From the top clusters we can see that our method effectively alleviates the salient region problem as the bounding boxes generated by our method better overlap with the ground truth boxes. 
Furthermore, from the bottom clusters, it is also evident that our method performs well when detecting multiple instances of the same class, which is often ignored by OICR + FRCNN \cite{tang2017multiple} detector.

There are two reasons behind the performance advantages of WSTDN. Firstly, our WSTDN backbone and BBM module have the ability to extract features that are not solely focused on the salient regions of the object. This allows the model to identify objects of a certain category based on their general and comprehensive features, even when dealing with multiple instances of the same class that have different gestures and appearances. 
Secondly, as illustrated in Figure~\ref{fig2}, many stages of refinements in the entire network generate many highest score proposals. Among these proposals, different instances of the same class may be selected as the highest score proposal at different refinement stages.

\subsection{Ablation Studies}

This section aims to investigate two key research questions through ablation studies. 
Firstly, we examine whether the WSTDN backbone outperforms the WSDDN in weakly supervised object detection. Secondly, we evaluate whether the collaborative work of BBM-based instance refinement and MTR modules enhances the our WSTDN backbone.

\paragraph{WSTDN vs WSDDN} \quad To compare the effectiveness of our proposed WSTDN backbone against the classic WSDDN \cite{bilen2016weakly}, we compare the two backbones with modules designed for WSDDN. As presented in Table~\ref{tab4}, we pair WSTDN with Online Instance Refinements (OIR) \cite{tang2017multiple}, ignoring lower IOU proposals (denoted as IOU$_{ign}$) \cite{2018PCL}, and bounding box regression (denoted as REG). Specifically, IOU$_{ign}$ means that, some proposals are ignored on purpose during training if its IOU with highest score proposal is lower than a threshold. This simple modification can avoid multiple instances of the same class from being classified to background to some extent. It is worth noting that as OIR, IOU$_{ign}$, and REG are all tailored for the classic WSDDN, this comparison naturally favours towards the classic backbone. However, from the results in Table~\ref{tab4} we can clearly see that our proposed WSTDN outperforms WSDDN in all circumstances. 
In all scenarios, WSTDN exhibits superior performance to WSDDN on both mAP and CarLoc, which illustrates the effectiveness of considering relationships between different proposals.

\begin{figure}[t]
\begin{center}
\subfigure[mAP]{
\includegraphics[width=0.45\linewidth]{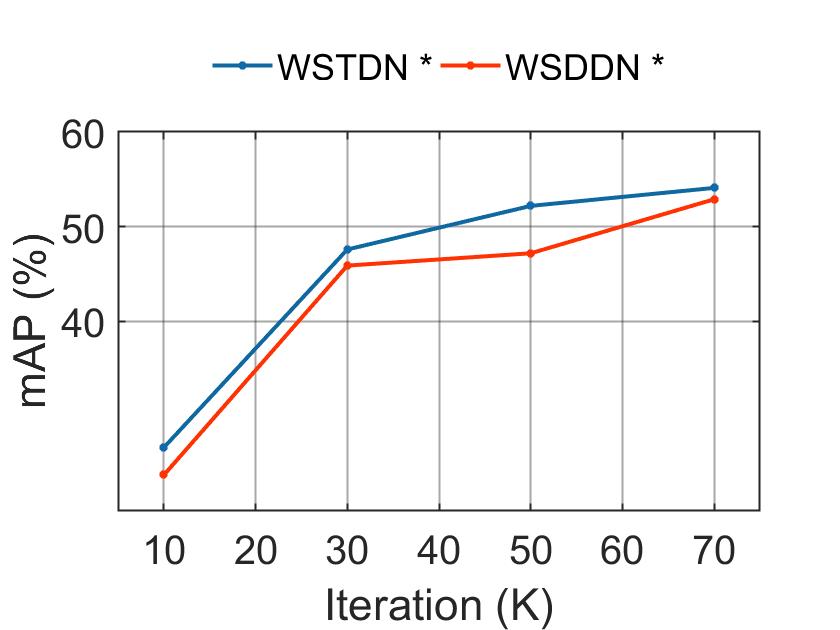}
}
\subfigure[CorLoc]{
\includegraphics[width=0.45\linewidth]{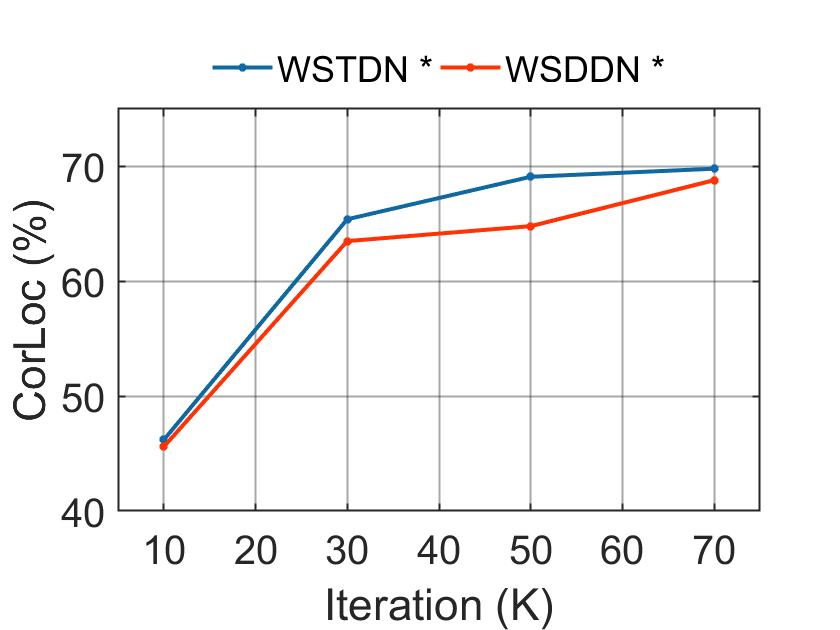}
}
\end{center}
\caption{Comparison of different baselines in training stages.}
\label{fig5}
\end{figure}

Furthermore, we also compare the detection performances of different training stages on VOC2007 to show the effectiveness of WSTDN. As illustrated in Figure~\ref{fig5}, where WSTDN$^*$ indicates using WSTDN with OIR + IOU$_{ign}$ + REG and WSDDN$^*$ indicates using WSDDN with the same modules, the results on mAP and CorLoc when trained for 10K, 30k, 50K, 70k iterations demonstrate that WSTDN outperforms WSDDN during the entire training process.

Lastly, we illustrate the invalidity of a naive implementation of the self-attention mechanism \cite{vaswani2017attention} instead of using our tailored ViT encoders. 
In Table~\ref{tab5}, Singe-SA + OIR indicates the model after replacing ViT encoders with single-headed self-attention, and Multi-Head-SA + OIR indicates the model after replacing ViT encoders to multi-headed self-attention. 
We can see that adding vanilla self-attention mechanisms actually significantly reduces performances. 
We attribute the reason as below. Firstly, in the context of Weakly Supervised Object Detection (WSOD), self-attention may encounter challenges in discovering relevant relationships among a large number of high-dimensional tokens (i.e. proposals).
Secondly, due to the absence of bounding box annotations, there exists an insufficiency of supervised information to guide the vanilla self-attention mechanism towards the optimal learning target.

\begin{table}[t]
\begin{center}
\small
    \begin{tabular}{|c|c|c|c|}
        \hline
        Baseline &components &mAP &CorLoc\\
        \hline
        &OIR  &41.2 &60.6\\
        WSDDN &OIR + IOU$_{ign}$  &42.2 &60.9\\
        &OIR + IOU$_{ign}$ + REG  &52.9 &68.8 \\
        \hline
        &OIR  &45.3 &63.0\\
        WSTDN &OIR + IOU$_{ign}$  &46.2 &62.9\\
        &\textbf{OIR + IOU$_{ign}$ + REG}  &\textbf{53.9} &\textbf{69.8} \\
        \hline
    \end{tabular}
\end{center}
\caption{Comparison between WSDDN and WSTDN baseline in terms of mAP (\%) and CorLoc (\%) on VOC2007 $test$ and $trainval$ set.}
\label{tab4}
\end{table}

\begin{table}[t]
\begin{center}
    \begin{tabular}{|c|c|}
        \hline
        Method & mAP(\%) \\
        \hline
        WSDDN + OIR  & 41.2 \\
        WSTDN + OIR &\textbf{45.3}\\
        Single-SA + OIR & 39.3\\
        Multi-Head-SA + OIR & 38.4 \\
        \hline
    \end{tabular}
\end{center}
\caption{ We utilize vanilla self attention mechanism\cite{vaswani2017attention} to replace ViT encoders, proving the invalidity of vanilla self attention\cite{vaswani2017attention} in WSOD. The experient is conducted on VOC2007 $test$ set.}
\label{tab5}
\end{table}

\begin{table}[t]
\begin{center}
\small
    \begin{tabular}{c|c|c|c|c}
        \hline
        WSTDN$^*$  & BBMR & MTR &mAP(\%) &CorLoc(\%)\\
        \hline\hline
        \checkmark  & & &53.9 &69.8\\
        \checkmark &\checkmark& &54.2 & 70.3\\
        \checkmark &\checkmark &\checkmark &\textbf{54.7} &\textbf{70.4} \\
        \hline
    \end{tabular}
\end{center}
\caption{Ablation study of different components of our method on VOC2007. WSTDN$^*$ indicates WSTDN adding OIR + IOU$_{ign}$ + REG, BBMR indicates BBM-based Refinements.}
\label{tab6}
\end{table}

\begin{figure}[t]
\begin{center}
\subfigure{
\includegraphics[width=0.25\linewidth]{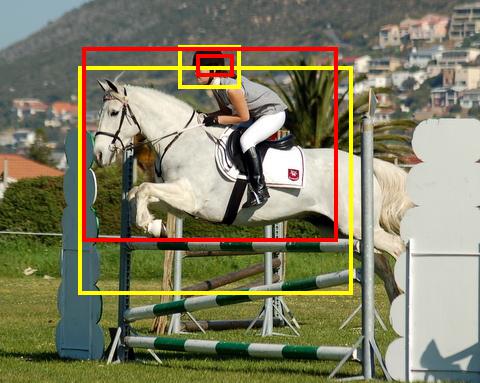}
}
\subfigure{
\includegraphics[width=0.27\linewidth]{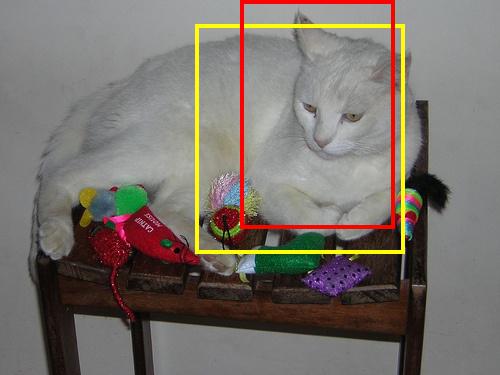}
}
\subfigure{
\includegraphics[width=0.27\linewidth]{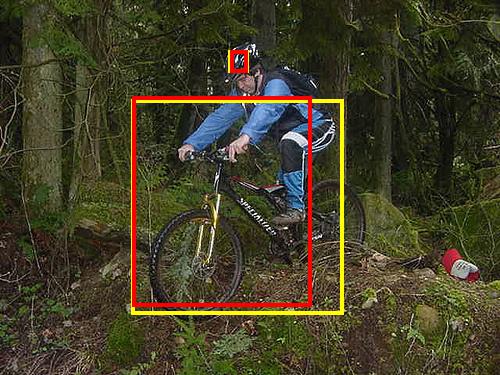}
}
\end{center}
\caption{Visualized results of BBM. Red boxes indicates the input of BBM, yellow boxes indicate the output of BBM.}
\label{fig6}
\end{figure}

\paragraph{The Utility of BBM and MTR}
Using WSTDN$^*$ as the initial model, we study the effectiveness of BBM-based instance refinements and the MTR modules. 
As shown in Table~\ref{tab6}, with only BBM-based instance refinements, we obtain a 0.3\% increase on mAP and 0.5 \% increase on CorLoc. 
After adding both BBM-based refinements and MTR, we obtain a further 0.8\% increase on mAP and 0.6 \% increase on CorLoc, compared to WSTDN$^*$. 
The results above indicates that BBM successfully alleviates the salient region problem by considering potential spatial relations between proposals, and MTR utilizes memory information to further enhance the ability of WSTDN and BBM in finding relations between proposals.
Figure~\ref{fig6} shows the mined bounding box of BBM, red and yellow boxes indicate the input and output of BBM respectively. It can be seen that BBM includes more object parts than before, reserving salient region and introducing no backgrounds.

\section{Conclusion}

In this paper, we propose an end-to-end Transformer-based WSOD framework to address the limitations of existing approaches that build upon the classic WSDDN. Our proposed WSTDN utilizes iterated self-attention and residual blocks to find inter-instance relationships, which for the first time replaces the classic WSDDN backbone. Then, BBM synthesizes novel bounding boxes to further alleviate salient region problem, thanks to the spatial relationships of high score proposals learned by WSTDN. Finally, MTR enhances WSTDN and BBM by combining history and current proposals. Our method alleviates both salient region problem and same-class-multi-instance problem. 
Experiments on publicly available benchmarks demonstrate that the proposed
approach achieves superior or competitive performances for WSOD.

{\small
\bibliographystyle{ieee_fullname}
\bibliography{egbib}
}

\end{document}